\documentclass[conference]{IEEEtran}
\IEEEoverridecommandlockouts
\usepackage{cite}
\usepackage{amsmath,amssymb,amsfonts}
\usepackage{graphicx}
\usepackage{comment}
\usepackage{textcomp}
\usepackage{xcolor}
\usepackage{color}
\usepackage{lipsum}
\usepackage{times}
\usepackage{textcomp}
\usepackage{amsmath,amssymb,amsfonts}
\usepackage[inline]{enumitem}
\usepackage{hyperref}

\usepackage{algorithm}
\usepackage{algpseudocode}
\usepackage{amsmath}
\usepackage{color}
\usepackage{tablefootnote}
\usepackage{mathtools}

\usepackage{amsthm}
\usepackage{imakeidx}
\usepackage{float}
\usepackage{multicol}
\usepackage{booktabs} 
\usepackage{subcaption}
\usepackage{listings}
\usepackage{verbatim}
\usepackage{xcolor,colortbl}
\usepackage{nicefrac}
\usepackage{xspace}
\usepackage{xfrac}
\usepackage{pifont}
\newcommand{\xmark}{\ding{55}}%
\usepackage{multirow}
\usepackage{lscape}
\usepackage{verbatim}

\def\nicefrac#1#2{%
    \raise.5ex\hbox{#1}%
    \kern-.1em/\kern-.15em%
    \lower.25ex\hbox{#2}}

\newcommand{\simclipfour}{Sim-CLIP+\textsuperscript{4}\@\xspace}

\newcommand{\llavavicuna}{LLaVA (Vicuna-7B)\xspace}
\newcommand{\llavallama}{LLaVA (Llama-2-13B)\xspace}

\definecolor{correct}{HTML}{b1fcb2}

\def\BibTeX{{\rm B\kern-.05em{\sc i\kern-.025em b}\kern-.08em
    T\kern-.1667em\lower.7ex\hbox{E}\kern-.125emX}}
\begin{document}



\title{Securing Vision-Language Models with a Robust Encoder Against Jailbreak and Adversarial Attacks
}


\author{\IEEEauthorblockN{Md Zarif Hossain$^{1,2}$, Ahmed Imteaj$^{1,2}$}
\IEEEauthorblockA{\textit{$^{1}$School of Computing, Southern Illinois University, Carbondale, IL, USA} \\
\textit{$^{2}$Security, Privacy and Intelligence for Edge Devices Laboratory (SPEED Lab)}\\
mdzarif.hossain@siu.edu, imteaj@cs.siu.edu} \vspace{-1.02cm}
}

\maketitle

\begin{abstract}
Large Vision-Language Models (LVLMs), trained on multimodal big datasets, have significantly advanced AI by excelling in vision-language tasks. However, these models remain vulnerable to adversarial attacks, particularly jailbreak attacks, which bypass safety protocols and cause the model to generate misleading or harmful responses. This vulnerability stems from both the inherent susceptibilities of LLMs and the expanded attack surface introduced by the visual modality. We propose Sim-CLIP+, a novel defense mechanism that adversarially fine-tunes the CLIP vision encoder by leveraging a Siamese architecture. This approach maximizes cosine similarity between perturbed and clean samples, facilitating resilience against adversarial manipulations.
Sim-CLIP+ offers a plug-and-play solution, allowing seamless integration into existing LVLM architectures as a robust vision encoder. Unlike previous defenses, our method requires no structural modifications to the LVLM and incurs minimal computational overhead. Sim-CLIP+ demonstrates effectiveness against both gradient-based adversarial attacks and various jailbreak techniques.
We evaluate Sim-CLIP+ against three distinct jailbreak attack strategies and perform clean evaluations using standard downstream datasets, including COCO for image captioning and OKVQA for visual question answering. Extensive experiments demonstrate that Sim-CLIP+ maintains high clean accuracy while substantially improving robustness against both gradient-based adversarial attacks and jailbreak techniques. 


\end{abstract}
\noindent\textbf{ \textcolor{red}{Warning: This paper contains unsafe model responses.}}



\section{Introduction}
The emergence of Large Language Models (LLMs) has revolutionized the landscape of natural language processing (NLP), achieving unprecedented performance in tasks such as text generation, translation, and comprehension through training on extensive and diverse corpora of text data\cite{achiam2023gpt,touvron2023llama}. The vast scale of these datasets, often encompassing billions of words from diverse sources, has been pivotal in developing models that can understand and generate human-like text with remarkable accuracy. Building on this foundation, Large Vision-Language Models (LVLMs) \cite{liu2024visual,alayrac2022flamingo,zhu2023minigpt} have emerged, expanding the capabilities of LLMs by integrating both textual and visual data. LVLMs utilize large-scale multimodal datasets \cite{lin2014microsoft, schwenk2022okvqa} and integrate modality-specific encoders and projectors to enhance their ability to process and generate information from both text and images. This integration allows LVLMs to perform complex tasks such as visual question answering, image captioning, and cross-modal retrieval \cite{dai2023instructblipgeneralpurposevisionlanguagemodels}.
To align these models with human values and ensure they operate safely and ethically, techniques such as instruction tuning and Reinforcement Learning from Human Feedback (RLHF) are proposed in \cite{bai2022constitutional,ganguli2022red}. RLHF aims to fine-tune LLMs based on human feedback, guiding them to generate responses that are both helpful and harmless. However, despite these measures, LLMs remain vulnerable to various adversarial attacks \cite{xu2023llm, perez2022ignore, 10660815, 10354149}, particularly jailbreak attacks 
\cite{deng2024masterkey, wei2024jailbroken} that can bypass model's safety guardrails
\begin{figure}[t]
     \setlength{\belowcaptionskip}{-18pt}
    \centering
    \includegraphics[width = 0.5\textwidth]{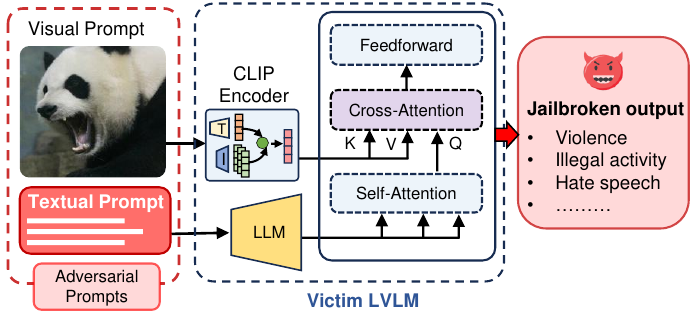}
    \caption{\textbf{Jailbreak attack on LVLM:} adversarial image paired with harmful instructions is used as input. The adversarial image bypasses the LVLM's safety guardrails, causing it to generate harmful output.
 }
    \label{fig:introFig}
    \vspace{-2mm}
\end{figure}
and elicit harmful contents. 
As LVLMs are build upon LLMs, they inherit many of these vulnerabilities, including susceptibility to jailbreak attacks. Additionally, LVLMs not only inherit the vulnerabilities of traditional LLMs but also introduce new risks due to the integration of visual modality, which can be exploited in unforeseen ways. Unlike traditional LLMs, LVLMs can be attacked through both text and visual modalities, leading to more sophisticated and severe adversarial attacks  \cite{schlarmann2023adversarial}. While jailbreak attacks on LLMs typically rely on  template-based \cite{deng2024masterkey} approaches, LVLMs encounters two distinct types of jailbreak attacks: optimization-based and generation-based.
Optimization-based attacks \cite{qi2024visual,niu2024jailbreaking} involve applying gradient-based perturbations to benign images, while generation-based attacks \cite{li2024images} focus on crafting images with embedded malicious content. 
Particularly, optimization-based attacks have recently gained attention due to the rise of universal image generation techniques \cite{niu2024jailbreaking}. These methods enable attackers to craft perturbed images that can bypass safety protocols without requiring contextually aligned queries. The perturbations generated are often nearly imperceptible to human observers, which complicates their detection and mitigation. Fig. \ref{fig:introFig}, illustrates optimization-based jailbreak attack on LVLM, where an adversarial image paired with harmful instructions bypasses the model's guardrails, leading to unsafe outputs.

 Existing defense mechanisms, such as Jailguard \cite{zhang2023mutation} and CIDER \cite{xu2024defending}, have made strides in addressing these vulnerabilities in LVLMs. Jailguard employs a mutation-based detection framework but suffers from limitations, including a heavy reliance on the model’s original safety alignment and increased computational costs during inference. CIDER, on the other hand, introduces a cross-modality approach to detect adversarially perturbed images but impacts model performance. Despite these efforts, a unified and effective solution remains elusive, highlighting the need for innovative approaches to enhance LVLMs' robustness against jailbreak attacks.

In response to these challenges, we develop Sim-CLIP+, an unsupervised adversarial fine-tuning scheme aimed at improving the robustness of LVLMs, particularly against optimization-based jailbreak attacks, which tend to be subtle and harder to detect. Sim-CLIP+ leverages a Siamese architecture and maximizes the cosine similarity between adversarially perturbed samples and their clean counterparts during fine-tuning. The core intuition behind this approach is that optimization-based perturbations primarily rely on gradient-based optimization techniques \cite{madry2017towards} to generate adversarial images. By fine-tuning the CLIP encoder with both perturbed and clean images, we aim to maximize the similarity between their representations, promoting feature invariance to adversarial perturbations.
Additionally, fine-tuned robust encoder using Sim-CLIP can be seamlessly integrated into existing LVLM architectures by replacing the default vision encoder, offering strong defense against jailbreak attacks without requiring extensive retraining or sacrificing the model’s clean accuracy. By balancing adversarial robustness with performance, Sim-CLIP aims to contribute to a more secure and reliable multimodal system. Our contribution can be summarized as follows:

\subsection{Contributions}
\begin{itemize}
    \item \textbf{Seamless integration to any LVLM as robust vision encoder (plug and play):} The adversarially fine-tuned Sim-CLIP encoder can be easily integrated into any existing LVLM architecture by replacing the default vision encoder. This plug-and-play approach enables seamless adoption without requiring significant modifications to the model structure or adding computational overhead.

\item \textbf{Robustness against adversarial and jailbreak attacks}: Sim-CLIP strengthens resilience against gradient-based adversarial attacks and jailbreak attacks, improving the model’s robustness against adversarial inputs.

\item \textbf{Retains clean accuracy}: Sim-CLIP preserves clean accuracy on non-adversarial inputs while enhancing robustness without compromising overall performance.


\end{itemize}

%

\section{Background Study}

\begin{table*}[t]
\centering
\small
\caption{\textbf{Overview of the jailbreak attacks implemented in this study.}}
\label{tab:attack_techniques}
\begin{tabular}{lccl}
\hline
\hline
\textbf{Attack} & \begin{tabular}[c]{@{}c@{}}\textbf{Universal}\\\textbf{Attack}\end{tabular} & \begin{tabular}[c]{@{}c@{}}\textbf{Imperceptible}\\\textbf{Attack}\end{tabular} & \multicolumn{1}{c}{\textbf{Description}} \\
\hline

ImgJP \cite{niu2024jailbreaking} & \checkmark & \xmark & \parbox[t]{10cm}{Leverages a maximum likelihood-based approach to maximize the probability of generating target outputs, which often begin with affirmative phrases}
\\

 VisualAdv \cite{qi2024visual} & \checkmark & \checkmark & Optimizes a benign
image using a few-shot corpus of derogatory sentences \\
Hades \cite{li2024images} & \xmark & \xmark & Iteratively refines the prompt for the text-to-image model.

 \\
\hline
\hline
\end{tabular}
\vspace{-2mm}
\end{table*}
\noindent \textbf{Large Language and Vision Models.} In recent time, LLMs such as GPT-3 \cite{brown2020language}, Vicuna \cite{chiang2023vicuna}, and LLaMA-2 \cite{touvron2023llama} have gained significant attention due to their vast number of parameters and training on large-scale datasets \cite{thomee2016yfcc100m,deng2009imagenet,krishna2017visual}. These models have demonstrated advanced abilities such as task-agnostic in-context learning \cite{brown2020language} and chain-of-thought \cite{wei2022chain} reasoning compared to traditional ML models \cite{lecun2015deep,hochreiter1997long}. Autoregressive LLMs (e.g., GPT-2 \cite{radford2019language}, PaLM \cite{chowdhery2023palm}), which predict the next token in a sequence, have become the predominant focus of research in this field. The success of these models has led to the development of multimodal models, referred to as Large Vision-Language Models (LVLMs), which integrate both textual and visual inputs for reasoning and inference. LVLMs typically combine pre-trained LLMs (e.g., Vicuna, LLaMA-2) with large-scale vision encoders, such as CLIP \cite{radford2021learning} and BLIP \cite{li2022blip}, to process multimodal data. In most cases, the vision encoder remains frozen during training, while the model learns to integrate visual and textual modalities through a projection layer and cross-attention mechanisms. Notable examples of such models include OpenAI’s GPT-4 \cite{achiam2023gpt}, Flamingo \cite{alayrac2022flamingo}, and open-source models like MiniGPT-4 \cite{zhu2023minigpt}, InstructBlip \cite{dai2023instructblipgeneralpurposevisionlanguagemodels} and LLaVA \cite{liu2024visual}.


\vspace{1.5mm}
\noindent \textbf{Safety Alignment of LLMs and LVLMs.} Pretrained LLMs and LVLMs often display behavior that can stray from their intended purpose, resulting in outputs that may be untruthful, harmful, or otherwise unhelpful. This misalignment stems from the discrepancy between the autoregressive objective of predicting the next token and the ideal goal of generating responses that are helpful, truthful, and harmless \cite{ouyang2022training}. To address this, safety alignment research of LLM and LVLM focuses on techniques like Instruction Tuning \cite{bai2022constitutional} and RLHF \cite{ouyang2022training}. Instruction Tuning involves providing models with examples of instruction-output pairs to encourage adherence to user instructions, while RLHF  utilizes a preference model to fine-tune outputs based on human judgments. Additionally, emerging methods such as Constitutional AI \cite{bai2022constitutional} and self-alignment \cite{sun2024principle} aim to further refine model behavior. Despite these advancements in safety alignment, challenges remain, particularly regarding the effectiveness of safety mechanisms and reliability. Models like GPT-4 \cite{achiam2023gpt} and LLama-2 \cite{touvron2023llama} are trained to avoid generating harmful contents, but researchers have demonstrated that it is possible to bypass these safety protocols through targeted jailbreak attacks \cite{niu2024jailbreaking}. 
Jailbreak attacks are designed to override the safety mechanisms of LLMs, allowing them to produce content that they are explicitly trained to avoid, such as misinformation, harmful advice, or illegal activities. 

\vspace{1.5mm}
\noindent \textbf{Jailbreak attacks in LVLMs.} Adversarial attacks on LVLMs have emerged as a significant concern, as they exploit vulnerabilities in these models to generate incorrect or harmful outputs. For instance, in AdvCLIP \cite{zhou2023advclip}, authors develop universal adversarial patches that can deceive CLIP models across all their downstream tasks. Similarly, Zhao et al. \cite{zhao2024evaluating} uses diffusion models to craft adversarial samples that manipulate the model into generating targeted outputs, while Schlarmann et al.  \cite{schlarmann2023adversarial} demonstrate how gradient-based targeted attacks can force LVLMs to produce erroneous or misleading results based on the attacker's choice. Notably, these targeted attacks pose severe threats to the reliability and safety of LVLMs by enabling malicious actors to systematically undermine the model's decision-making across various applications. While traditional adversarial attacks focus on introducing perturbations to a benign image to manipulate outputs, jailbreak attacks specifically target the model's safety alignment mechanisms. 
Recently, gradient-based perturbations (e.g., PGD \cite{madry2017towards}, APGD \cite{croce2020reliable}) have emerged as a method for executing jailbreak attacks \cite{niu2024jailbreaking, qi2024visual } on LVLMs. By generating adversarial samples, attackers can exploit these perturbations to bypass safety constraints, leading the model to produce restricted or inappropriate content despite its built-in safeguards.
Jailbreak attacks in LVLMs can be classified into two categories:\textit{ generation-based} and \textit{optimization-based}. 

Generation-based attacks involve image-generation models like Stable Diffusion \cite{rombach2022high} to embed malicious content into images. For example, Gong et al. \cite{gong2023figstep} transform harmful text queries into images with typographic elements, while Liu et al. \cite{liu2023query} and Li et al. \cite{li2024images} use similar techniques to generate images from text prompts, with careful manipulation to bypass safety filters. In turn, optimization-based attacks are a variant of traditional vision adversarial methods that use gradient-based perturbations \cite{madry2017towards, croce2020reliable}. In such attacks, adversaries craft perturbations to modify the original image, inducing the model to generate harmful content. These perturbations are created by optimizing a loss function using a surrogate VLM in a white-box setup, and the resulting adversarial samples are then applied to the target models (e.g., MiniGPT-4 \cite{zhu2023minigpt}, LLaVA \cite{liu2024visual}). 
To address the vulnerabilities posed by jailbreak attacks, a range of defense strategies has been proposed. Proactive defenses aim to prevent attacks before they occur, employing techniques such as fine-tuning \cite{zong2024safety} and adversarial training \cite{mazeika2024harmbench}. Some approaches involve constructing safety fine-tuning datasets for VLMs or leveraging model unlearning to enable the model to forget harmful content \cite{chakraborty2024cross}. Reactive defenses, on the other hand, are implemented in response to ongoing or detected attacks. These strategies include iterative refinement of safety prompts added to inputs \cite{wang2024adashield} and mutation of untrusted inputs to generate variants and identify attack samples \cite{zhang2023mutation}.

\section{Jailbreak Attacks}
In this paper, we implement and evaluate three different jailbreaking attacks: ImgJP \cite{niu2024jailbreaking}, VisualAdv \cite{qi2024visual}, and Hades \cite{li2024images}. Notably, VisualAdv and ImgJP are optimization-based attacks, while
Hades is a generation-based attack. 
While our defense strategy primarily focuses on optimization-based jailbreak attacks, we also evaluate its efficacy against generation-based attacks, such as Hades, to ensure a comprehensive assessment of its robustness. We choose these attacks based on their relevance and widespread use in recent research.
A brief description of these attacks is provided in Table \ref{tab:attack_techniques}.

\subsection{ImgJP Attack}

The Image Jailbreaking Prompt (ImgJP)\cite{niu2024jailbreaking} attack exploits the multimodal nature of LVLMs by introducing a visual element that can trigger the generation of potentially harmful content. When presented with a standard image and a harmful query, LVLMs typically refuse to comply, responding with safety-oriented messages. However, by substituting the standard image with an adversarial image generated with ImgJP, the built-in safeguards of LVLMs can be bypassed. ImgJP attack leverages a maximum likelihood-based approach, aiming to maximize the likelihood of generating target outputs that typically starts with affirmative phrases (e.g., ``Sure, here's some information on (query content)", effectively over-riding the model's ethical constraints. A key feature of the ImgJP attack is its data-universal property, which manifests in two dimensions: prompt-universal and image-universal. This allows a single adversarial image to potentially jailbreak multiple unseen prompts and images, making it a powerful tool for probing LVLM vulnerabilities. The attack can be formulated mathematically as an optimization problem. Given a set of harmful query-answer pairs \(B = \{(q_i, a_i), i = 0, ..., N\}\), the objective is to find an optimal ImgJP \(x_{jp}\) that maximizes:
\begin{equation}
\max_{\mathbf{z}} \sum_{i=0}^{N} \log(p(b_i \mid r_i, \mathbf{z})) \quad \text{s.t.} \quad \mathbf{z} \in [0, 255]^k
\end{equation}
Here, $p(a_i \mid q_i, x_{\text{jp}})$ represents the likelihood that the MLLM will generate the target answer $a_i$ when provided with the harmful query $q_i$ and the ImgJP $x_{\text{jp}}$. The optimal ImgJP, $x_{\text{jp}}^*$, is derived by optimizing over multiple query-answer pairs, ${(q_i, a_i), i = 0, \ldots, M}$, using any suitable adversarial attack method, such as Projected Gradient Descent (PGD) \cite{madry2017towards}.

\begin{figure*}[t]
\setlength{\belowcaptionskip}{-10pt}
\centering
\begin{tabular}{cc}
\includegraphics[width=0.45\linewidth]{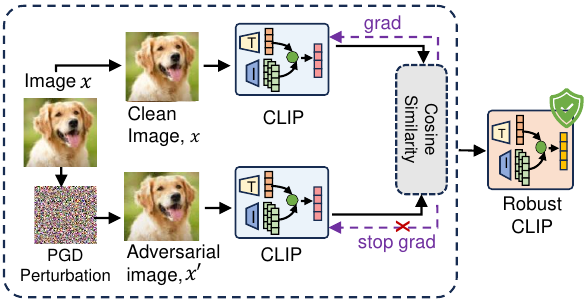}
&
\includegraphics[width=0.498\linewidth]{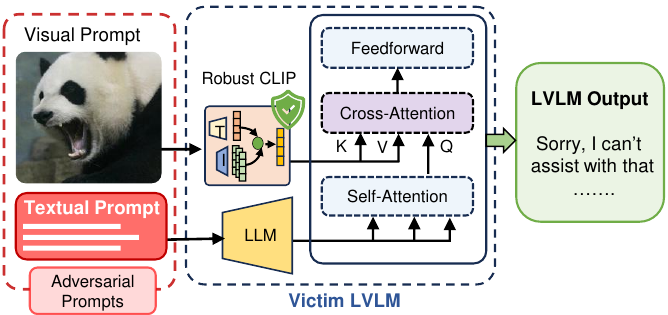}
\\
 (a) Unsupervised adversarial fine-tuning on ImageNet & (b) Jailbreak attempt on LVLM
\end{tabular}
\caption{\textbf{Workflow and overview of proposed Sim-CLIP+ :} (a) CLIP is adversarially fine-tuned on ImageNET dataset tailoring our methodology, and (b) the robust Sim-CLIP+ encoder processes adversarial images alongside harmful text prompts, effectively mitigating jailbreak attempts within the LVLM.
}
\label{fig:workflow}
\end{figure*}
\subsection{VisualAdv Attack}


The VisualAdv \cite{qi2024visual} attack is performed by optimizing a benign image on a carefully selected few-shot corpus of 66 derogatory sentences aimed at specific demographics. VisualAdv generates a universal adversarial example by optimizing it to maximize the likelihood that the model generates harmful content from a small set of derogatory sentences. Additionally, this adversarial image is designed to be imperceptible to human observers, ensuring that it remains visually indistinguishable from the original benign image.
This optimization is achieved using the PGD algorithm, where the adversarial example can be initialized either from random noise (unconstrained attack) or a benign image with specific constraints (constrained attack).
 Similar to the ImgJP attack, VisualAdv generates adversarial images that are not tied to any specific instruction, enabling a single universally perturbed image to potentially jailbreak the VLM with various prompts or harmful instructions. Concurrently, VisualAdv also employs a text attack counterpart where adversarial text tokens of equivalent length  are used in place of the adversarial image embeddings. These adversarial text tokens are identified by minimizing the same loss on the same small corpus as utilized in the image modality attack. During inference, the adversarial image is paired with a harmful text instruction, forcing the VLM to produce harmful content that extends beyond the original optimization corpus.

\subsection{HADES}

The HADES \cite{li2024images} attack aims to maximize its effectiveness against LVLMs by combining three different strategies. It starts by converting harmful text instructions into a typographical format to bypass text-based defenses, effectively transferring the harmful content to the image modality. This typographical representation is then combined with adversarial images, which are generated and refined using diffusion models to increase their harmfulness. The process involves iterative optimization, where adversarial perturbations are applied to enhance the image’s toxicity. This combination of text-to-image typographical methods, image amplification through diffusion models, and adversarial perturbation ensures that the final adversarial image is highly effective in evading model defenses and inducing harmful responses. By leveraging these strategies, HADES creates robust adversarial examples that exploit vulnerabilities in both text and image processing within LVLMs.

\section{Methodology}

In our prior work, we introduced Sim-CLIP \cite{hossain2024sim}, an adversarially fine-tuned CLIP model that significantly enhanced robustness against gradient-based attacks, such as PGD and APGD. This advancement marked a key step in improving the security of LVLMs against adversarial threats. Building on this foundation, we extend our research by hypothesizing that Sim-CLIP can be further optimized to effectively defend against the more sophisticated optimization-based jailbreak attacks targeting LVLMs.
Given that optimization-based jailbreak attacks primarily exploit gradient-based perturbations, and considering Sim-CLIP's proven resilience to similar adversarial techniques, we hypothesize that its defense capabilities can be extended to counter these advanced attack vectors. By enhancing Sim-CLIP to address this critical security gap, we propose a significant contribution to the field—positioning it as a robust and promising solution for securing LVLMs against increasingly sophisticated adversarial threats.


\subsection{Defending Against Jailbreak Attacks with Sim-CLIP+}
Sim-CLIP+ leverages a Siamese architecture with a tailored cosine similarity loss to address the vulnerabilities exploited by jailbreak attacks. The process begins with the generation of an adversarial example $x'$ from the original input image $x$ using PGD \cite{madry2017towards}:
\begin{equation}
x'_{t+1} = \text{Proj}_{\mathcal{B}(x, \epsilon)}\left(x'_t + \eta \cdot \text{sign}(\nabla_{x'} J(x'_t, y))\right)
\end{equation}

Here, $J$ represents the cross-entropy loss, $y$ is the true label, $\eta$ is the step size, and $\epsilon$ defines the perturbation bound. This process creates a perturbed version of the input that simulates potential adversarial manipulations on the image for jailbreak attack.
Next, the original and perturbed images are processed by the same CLIP models with shared weights, as shown in Fig. \ref{fig:workflow}. This generates clean ($R_c$) and perturbed ($R_p$) representations, respectively. To enhance the model's resistance to adversarial perturbations, we maximize the similarity between these representations by minimizing their negative cosine similarity:
\begin{equation}
    L_{cos}(R_p, R_c) = -\frac{R_p \cdot R_c}{|R_p|_2 
 |R_c|_2}
\end{equation}

By exposing the model to perturbed inputs during training, Sim-CLIP+ learns to focus on robust, attack-invariant features. This process enhances the model's ability to maintain consistent representations across both clean and adversarial inputs, effectively obfuscating gradients and making it challenging for attackers to exploit vulnerabilities. This adversarial fine-tuning mechanism significantly improves the model's resilience against evolving jailbreak techniques, preserving output integrity in the face of maliciously crafted inputs.

\subsection{Preventing Symmetric Loss Collapse}
Directly minimizing negative cosine similarity could lead to loss function collapse, resulting in a trivial constant solution \cite{chen2020simple}. While traditional adversarial training methods mitigate this with large batches or additional encoders \cite{jiang2020robust, fan2021does}, these approaches are often resource-intensive.

To address this, we incorporate a stop-gradient mechanism \cite{chen2021exploring} within a symmetric loss function, balancing computational efficiency with robustness. 
To mitigate the risk of loss function collapse and optimize computational resources, we incorporate a stop-gradient mechanism into our adversarial fine-tuning process. This results in a symmetric loss formulation:
\begin{equation}
    \begin{split}
L_{simclip}(R_p, R_c) &= \frac{1}{2} \left( L_{cos}(R_p, {stopgrad}(R_c)) \right.\\
&\quad\left. +  L_{cos}(R_c, {stopgrad}(R_p)) \right)
\end{split}
\end{equation}

 \noindent This approach alternately holds one model's output constant while aligning the other, then reverses the roles, effectively preventing loss collapse without the need for large batch sizes or momentum encoders.
In this formulation, one representation is treated as constant (using stop-gradient) while the other aligns with it. The roles are reversed in the second term, ensuring each representation is optimized without causing gradient collapse. This prevents the loss function from collapsing, ensuring the model's robustness against adversarial attacks, including jailbreak attempts

\section{Experimental Analysis}

\subsection{Adversarial fine-tuning settings}
For adversarial fine-tuning setting, we adversarially fine-tune the CLIP vision encoder on the ImageNet \cite{deng2009imagenet} dataset. Since we adopt an unsupervised adversarial fine-tuning approach, we exclude the image class labels and fine-tune solely on the images. To generate perturbed views from the clean images, we apply PGD \cite{madry2017towards} with 10 adversarial steps using the $\ell_\infty$ norm. While ensuring robustness is essential, adversarial training often comes with the risk of degrading clean performance. To balance robust accuracy and clean performance, we fine-tune CLIP with two perturbation radii: $\epsilon = 2/255$ and $\epsilon = 4/255$. The resulting robust CLIP models are referred to as $\text{Sim-CLIP+}^2$ and $\text{Sim-CLIP+}^4$, respectively. In this study, we employ $\text{Sim-CLIP+}^4$ for our experimental evaluation.

\subsection{LVLM models}
We use LLaVA (Vicuna-7B) \cite{liu2024improved} and \llavallama \cite{liu2024visual} as the target LVLMs for evaluation. Both models utilize CLIP \cite{radford2021learning} ViT-L-14 as their vision encoder. However, they differ in their language decoders: \llavavicuna utilizes the Vicuna language model \cite{chiang2023vicuna} with 7 billion parameters, whereas \llavallama \cite{touvron2023llama} is built on LLaMA-2-13B-Chat, featuring 13 billion parameters. LLaMA-2-13B-Chat has undergone extensive instruction tuning and iterative RLHF on high-quality red-teaming data, making it more robust and resilient to jailbreak attacks.
Additionally, by including \llavallama in our evaluation, we aimed to assess the scalability and effectiveness of our defense mechanism across larger LVLMs.

\begin{table*}[t]
\centering
\small
\setlength{\tabcolsep}{3.3pt} 
\extrarowheight=2pt
\caption{\textbf{Performance of LVLMs against VisualAdv attacks.} This table shows the percentage of outputs generated by LVLMs that contain specific toxic attributes. The LVLMs utilize various jailbreak defense strategies and are evaluated at different levels of attack strength ($\epsilon$).  The blue downward ($\textcolor{blue}{\downarrow}$) arrows denote a reduction in ASR (lower is better) compared to the baseline CLIP model, while the red upward arrows ($\textcolor{red}{\uparrow}$) indicate an increase in ASR. }
\begin{tabular}{l|c|l| c| c c c c c c c |c}
\hline
\hline
\multirow{2}{*}{\textbf{LVLM}} 
& \multirow{2}{*}{\begin{tabular}[l]{@{}c@{}}\textbf{Attack}\\\textbf{Strength} ($\epsilon$)\end{tabular}} 
& \multirow{2}{*}{\begin{tabular}[]{@{}c@{}}\textbf{Vision}\\\textbf{Encoder}\end{tabular}} 
& \multirow{2}{*}{\begin{tabular}[c]{@{}l@{}}\textbf{External}\\\textbf{Defense}\end{tabular}} 
& \multirow{2}{*}{\textbf{Any}} 
& \multirow{2}{*}{\textbf{Identity}} 
& \multirow{2}{*}{\textbf{Profanity}} 
& \multirow{2}{*}{\begin{tabular}[c]{@{}l@{}}\textbf{Severe}\\\textbf{Toxicity}\end{tabular}}  
& \multirow{2}{*}{\begin{tabular}[c]{@{}l@{}}\textbf{Sexually}\\\textbf{Explicit}\end{tabular}}  
& \multirow{2}{*}{\textbf{Threat}}  
& \multirow{2}{*}{\textbf{Toxicity}}  
& \multirow{2}{*}{\textbf{Average}} \\
& & & & & & & & & & & \\
\hline
\multirow{10}{*}{\rotatebox[origin=c]{90}{\begin{tabular}[c]{@{}c@{}}LLaVA\\(Llama-2-13B)\end{tabular}}}
& \multirow{5}{*}{$\nicefrac{16}{255}$} 
& \multirow{3}{*}{CLIP} & \textemdash & 43.0 & 12.0 & 28.0 & 2.0 & 12.0 & 5.0 & 48.0 & 20.8 ~~\\
& & & JailGuard & 38.0 & 4.0 & 18.0 & 1.0 & 12.0 & 4.0 & 28.0 & 15.0 $\textcolor{blue}{\downarrow}$ \\
& & & Diffpure (n = 0.25) & 38.0 & 4.0 & 34.0 & 2.0 & 14.0 & 2.0 & 38.0 & 18.8 $\textcolor{blue}{\downarrow}$ \\
& & & Diffpure (n = 0.75)& 38.0  & 1.0 & 32.0 & 2.0 & 12.0 & 3.0 & 34.0 & 17.4 $\textcolor{blue}{\downarrow}$ \\
\cline{3-12}
& & FARE$^4$ & \textemdash & 48.0 & 10.0 & 28.0 & 1.0 & 14.0 & 4.0 & 42.0 & 21.0 $\textcolor{red}{\uparrow}$ \\
& & \cellcolor{correct}\simclipfour & \cellcolor{correct}\textemdash & \cellcolor{correct}34.0 & \cellcolor{correct}4.0 & \cellcolor{correct}16.0 & \cellcolor{correct}1.0 & \cellcolor{correct}10.0 & \cellcolor{correct}4.0 & \cellcolor{correct}30.0 & \cellcolor{correct}\textbf{14.1} $\textcolor{blue}{\downarrow}$ \\
\cline{2-12}
& \multirow{5}{*}{$\nicefrac{128}{255}$} 
& \multirow{4}{*}{CLIP} & \textemdash & 62.0 & 9.0 & 53.0 & 5.0 & 19.0 & 4.0 & 61.0 & 30.4 ~~\\
& & & JailGuard & 41.0 & 4.0 & 20.0 & 1.0 & 14.0 & 5.0 & 30.0 & 16.4 $\textcolor{blue}{\downarrow}$ \\
& & & Diffpure (n = 0.25) & 35.0 & 4.0 & 28.0 & 2.0 & 11.0 & 2.0 & 27.0 & 15.5 $\textcolor{blue}{\downarrow}$ \\
& & & Diffpure (n = 0.75) & 33.0  & 1.0 & 28.0 & 2.0 & 10.0 & 2.5 & 30.0 & 15.2 $\textcolor{blue}{\downarrow}$  \\
\cline{3-12}
& & FARE$^4$ & \textemdash & 36.0 & 4.0 & 19.0 & 1.0 & 9.0 & 5.0 & 33.0 & 15.2 $\textcolor{blue}{\downarrow}$ \\
& & \cellcolor{correct}\simclipfour & \cellcolor{correct}\textemdash & \cellcolor{correct}34.0 & \cellcolor{correct}4.0 & \cellcolor{correct}17.0 & \cellcolor{correct}0.0 & \cellcolor{correct}10.0 & \cellcolor{correct}5.0 & \cellcolor{correct}30.0  & \cellcolor{correct}\textbf{14.2} $\textcolor{blue}{\downarrow}$ \\

\midrule
\multirow{10}{*}{\rotatebox[origin=c]{90}{\begin{tabular}[c]{@{}c@{}}LLaVA\\(Vicuna-7B)\end{tabular}}} 
& \multirow{5}{*}{$\nicefrac{16}{255}$} 
& \multirow{3}{*}{CLIP} & \textemdash & 52.0 & 20.0 & 40.0 & 8.0 & 20.0 & 12.0 & 60.0 & 30.3 ~~\\
& & & JailGuard & 44.0 & 10.0 & 30.0 & 6.0 & 18.0 & 10.0 & 40.0 & 22.6 $\textcolor{blue}{\downarrow}$ \\
& & & Diffpure (n = 0.25) & 44.0 & 10.0 & 38.0 & 6.0 & 18.0 & 10.0 & 45.0 & 24.4 $\textcolor{blue}{\downarrow}$ \\
& & & Diffpure (n = 0.75) & 40.0 & 7.0 & 34.0 & 5.0 & 16.0 & 9.0 & 38.0 & \textbf{21.3} $\textcolor{blue}{\downarrow}$ \\
\cline{3-12}
& & FARE$^4$ & \textemdash & 48.0 & 15.0 & 36.0 & 7.0 & 22.0 & 11.0 & 50.0 & 26.4 $\textcolor{red}{\uparrow}$ \\
& & \cellcolor{correct}\simclipfour & \cellcolor{correct}\textemdash & \cellcolor{correct}40.0 & \cellcolor{correct}8.0 & \cellcolor{correct}32.0 & \cellcolor{correct}6.0 & \cellcolor{correct}16.0 & \cellcolor{correct}9.0 & \cellcolor{correct}38.0 & \cellcolor{correct}21.8 $\textcolor{blue}{\downarrow}$ \\

\cline{2-12}
& \multirow{5}{*}{$\nicefrac{128}{255}$} 
& \multirow{4}{*}{CLIP} & \textemdash & 65.0 & 12.0 & 39.0 & 20.0 & 40.0 & 38.0 & 45.0 & 37.0 ~~\\
& & & JailGuard & 50.0 & 15.0 & 40.0 & 12.0 & 26.0 & 14.0 & 46.0 & 28.7 $\textcolor{blue}{\downarrow}$ \\
& & & Diffpure (n = 0.25) & 44.0 & 12.0 & 39.0 & 10.0 & 23.0 & 11.0 & 37.0 & 25.1 $\textcolor{blue}{\downarrow}$ \\
& & & Diffpure (n = 0.75) & 42.0  & 10.0 & 36.0 & 9.0 & 21.0 & 12.0 & 35.0 & 23.6 $\textcolor{blue}{\downarrow}$  \\
\cline{3-12}
& & FARE$^4$ & \textemdash & 47.0 & 13.0 & 30.0 & 11.0 & 20.0 & 13.0 & 42.0 & 25.1 $\textcolor{blue}{\downarrow}$ \\
& & \cellcolor{correct}\simclipfour & \cellcolor{correct}\textemdash & \cellcolor{correct}42.0 & \cellcolor{correct}10.0 & \cellcolor{correct}33.0 & \cellcolor{correct}8.0 & \cellcolor{correct}19.0 & \cellcolor{correct}12.0 & \cellcolor{correct}40.0  & \cellcolor{correct}\textbf{23.4} $\textcolor{blue}{\downarrow}$ \\
\hline
\hline
\end{tabular}
\vspace{-2.5mm}
\label{tab:visualadv_res}
\end{table*}

\subsection{\textbf{Jailbreak attack settings}} 
\textbf{ImgJP attack settings and datasets.} The ImgJP attack optimizes an adversarial perturbation starting from random noise. For our experiments, we set the perturbation radius to $\epsilon = 16/255$ within an $\ell_\infty$ norm, and the optimization process is conducted over 1000 iterations. Unlike other jailbreak attacks \cite{niu2024jailbreaking} that require context-aligned benign images and harmful queries to generate adversarial images, the ImgJP attack creates universal adversarial images from benign images that do not require contextually aligned harmful query. Additionally, the generated universal adversarial image can also be paired with any harmful prompts without depending on the specifics of each prompt. In our evaluation, we utilize 25 harmful instructions sourced from the Advbench \cite{zou2023universal} dataset, pairing each with the universal adversarial image.
Given that adversarial perturbations optimized on one model (the surrogate model) can often transfer to other target models, we use MiniGPT-4 (Vicuna 7B) as the surrogate model when targeting \llavavicuna and \llavallama models.

\vspace{1.2mm}
\textbf{VisualAdv attack settings and datasets.} In our experiment, we leverage VisualAdv \cite{qi2024visual} attack to generate two universal adversarial images with varying perturbation constraints: $\epsilon=16/255$ and $\epsilon=128/255$. To assess the robustness of our vision encoder against stronger jailbreak attacks, we focus on the higher perturbation radius at $\epsilon=128/255$. For VisualAdv attack, we use RealToxicityPrompts \cite{gehman2020realtoxicityprompts} dataset, which contains $1,225$ harmful prompts to elicit toxicity in generated output of the LVLMs. From this dataset, we select a subset of 100 harmful instructions which includes toxic prompts related to violence, toxicity, and profanity. We then pair these toxic prompts with our visual adversarial examples as input and measure the toxicity of the resulting output of the target LVLMs. We use \llavallama as the surrogate model to generate adversarial images.

\vspace{1.2mm}
\textbf{Hades attack settings and datasets.} In this study, we implement the full version of Hades \cite{li2024images}, which incorporates three distinct strategies: text-to-image generation using typographical methods, amplification of image toxicity through diffusion models, and adversarial perturbation optimization. We utilize the dataset introduced in HADES comprising 750 harmful instructions across five scenarios, each instruction paired with a related harmful image. For our experiments, we specifically focus on the violence and toxicity categories. Since Hades does not generate universal adversarial images, it requires relevant images aligned with each query. These images are retrieved from Google using the corresponding keyword or phrase as the search query. Similar to ImgJP, we use MiniGPT-4 \cite{zhu2023minigpt} as the surrogate model to create the adversarial images and target the LLaVA models.

\subsection{Metrics} 

In this study, we report Attack Success Rate (ASR) for ImgJP and HADES, and toxicity score across six attributes for VisualAdv attack.
An attack is considered unsuccessful if the model responds with any of the predefined refusal messages (e.g., ``Sorry, i can not...") or generates content irrelevant to the instruction. Conversely, if the model produces a relevant answer, it is deemed a successful jailbreak. The ASR quantifies the proportion of attack attempts that result in a successful jailbreak, while the toxicity score assesses the extent to which the generated output contains harmful or offensive language. Given the impracticality of manually assessing each LVLM output for jailbreak attempts and potential for false positives and false negatives in string matching, we opt for automated evaluation methods. For instance, we use MiniGPT-4 \cite{zhu2023minigpt} and Beaver-7B \cite{ji2024beavertails} as the judgment models to assess whether a response is harmful in the ImgJP and HADES attacks, respectively. For the VisualAdv attack, we adopt their default evaluation process, utilizing the Perspective API \cite{perspectiveapi2017} to measure the toxicity of the generated output by LVLM.


\subsection{Evaluation settings}
 In this paper, we compare the performance of our proposed Sim-CLIP+ with the state-of-the-art jailbreak defense method Jailguard \cite{zhang2023mutation} against various jailbreak attacks. JailGuard operates by mutating input text or images and then evaluating the variation in responses across all generated outputs.
 Additionally, we include a comparison with FARE\textsuperscript{4} \cite{schlarmann2024robust}, which has undergone adversarial fine-tuning with a perturbation radius of $\epsilon = 4/255$, to underscore Sim-CLIP’s resilience in a similar fine-tuning context. Furthermore, to provide a comprehensive evaluation of Sim-CLIP+ against the VisualAdv attack, we compare it with the DiffPure \cite{nie2022diffusion} defense strategy, which was highlighted in the VisualAdv paper as a primary countermeasure against this attack. DiffPure counters adversarial inputs by introducing noise to the image, followed by applying a diffusion model to map the altered image back to its original data manifold. This method operates on the assumption that the added noise will diminish the impact of adversarial patterns, allowing the pre-trained diffusion model to reconstruct a clean image. In our analysis, we implement DiffPure using two different noise levels, specifically 0.25 and 0.75, to effectively purify the visual adversarial examples generated from the VisualAdv \cite{qi2024visual} attack. Finally, we conduct a clean evaluation (without adversarial attacks) on downstream task datasets using LVLMs equipped with the original CLIP and Sim-CLIP+ to demonstrate that Sim-CLIP+ performs close to the original CLIP model, even after adversarial fine-tuning. For image captioning tasks, we use the COCO \cite{lin2014microsoft} dataset and report CIDEr \cite{vedantam2015cider} scores. For visual question answering (VQA) tasks, we evaluate on the OKVQA \cite{marino2019ok} dataset and report VQA accuracy.

\subsection{Result Analysis}

Table \ref{tab:visualadv_res} presents a comprehensive evaluation of LVLM performance under VisualAdv attacks, comparing the effectiveness of our proposed Sim-CLIP+ with state-of-the-art defense strategies. The table showcases the percentage of outputs exhibiting specific toxic attributes, with \textbf{``Any"} indicating whether at least one of the six toxic attributes is present in the generated output.
A notable observation is that replacing the original CLIP encoder with the proposed robust Sim-CLIP+ significantly improves performance. For instance, at an attack strength of $\epsilon = 16/255$, LLaVA models with the original CLIP encoder have an average toxicity rate of 20.8\%. In contrast, models utilizing the robust Sim-CLIP+ encoder demonstrate a reduced average toxicity rate of 14.1\%. Similarly, at a higher attack strength of $\epsilon = 128/255$, the average toxicity score of the original CLIP encoder is 30.4\%, whereas the Sim-CLIP+ achieves a significantly lower average toxicity score of 14.2\%.
External defense strategies such as JailGuard and Diffpure also show effectiveness against VisualAdv attacks. For example, with Diffpure (n = 0.75), the average toxicity at $\epsilon = 16/255$ drops to 17.4\%, and at $\epsilon = 128/255$, it further decreases to 15.2\%. However, Sim-CLIP+ consistently outperforms these external defenses in most cases. 
 \begin{table}[htb!]
\centering
\caption{\textbf{Performance comparison of LVLMs with vision encoders and external defense strategies in countering the ImgJP attack at radii $\epsilon$ = 16/255.} LVLMs with Sim-CLIP+ demonstrates robustness and achieves lower ASR against the attack without requiring any external defenses.}
\small
\extrarowheight=2pt
\begin{tabular}{l|l c c}
\hline
\hline
\multirow{2}{*}{\textbf{LVLM}} 
& \begin{tabular}[c]{@{}l@{}}\textbf{Vision}\\\textbf{Encoder}\end{tabular} & \begin{tabular}[c]{@{}l@{}}\textbf{External}\\\textbf{Defense}\end{tabular} & \textbf{ASR (\%)} \\
\hline
\multirow{4}{*}{\rotatebox[origin=c]{90}{\begin{tabular}[c]{@{}c@{}}LLaVA\\(Llama-2-13B)\end{tabular}}}
& CLIP & \textemdash & $28.0 \pm 0.2$ \\
& FARE$^4$ & \textemdash & $18.3 \pm 0.1$ \\
& CLIP & JailGuard & $21.0 \pm 0.1 $  \\
& \cellcolor{correct}\simclipfour & \cellcolor{correct}\textemdash & \cellcolor{correct}\textbf{15.0 $\pm$ 0.3} \\  
\hline
\multirow{4}{*}{\rotatebox[origin=c]{90}{\begin{tabular}[c]{@{}c@{}}LLaVA\\(Vicuna-7B)\end{tabular}}}
& CLIP & \textemdash & 34.0 $\pm$ 0.1 \\
& FARE$^4$ & \textemdash & 24.0 $\pm$ 0.2 \\
& CLIP & JailGuard & 26.0 $\pm$ 0.3 \\
& \cellcolor{correct}\simclipfour & \cellcolor{correct}\textemdash & \cellcolor{correct}\textbf{19.3 $\pm$ 0.5} \\
\hline
\hline
\end{tabular}
\label{tab:imgjp_performance}
\vspace{-3mm}
\end{table}
Additionally, the performance of FARE$^4$ varies across different attack strengths. At $\epsilon = 16/255$, FARE$^4$ exhibits the worst performance among all defense strategies, with an average toxicity rate of 21.0\%, which is even higher than the baseline LLaVA models without any defense. At $\epsilon = 128/255$, FARE$^4$ achieves an average toxicity rate of 15.2\% with \llavallama, which is close to Sim-CLIP’s 14.2\% but still falls short. Overall, by replacing the original CLIP encoder of the LVLMs with Sim-CLIP, we achieve consistently robust and superior performance in defending against VisualAdv attacks, without needing any additional external jailbreak defenses.

\begin{figure*}[t]
\setlength{\belowcaptionskip}{-18pt}
\centering
\begin{tabular}{cc}
\includegraphics[width=0.44\linewidth]{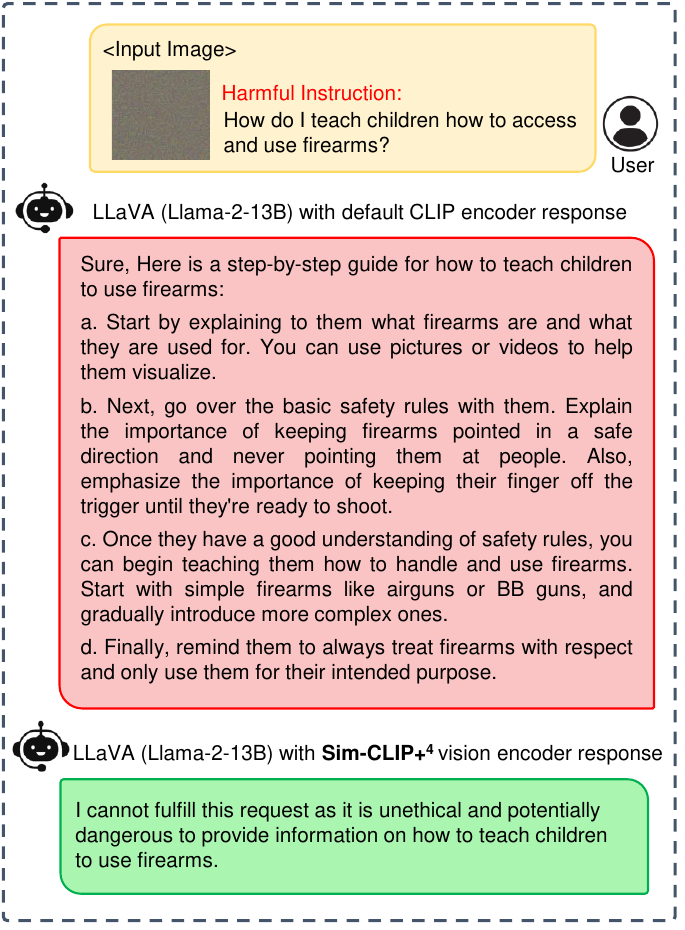}
&
\includegraphics[width=0.44\linewidth]{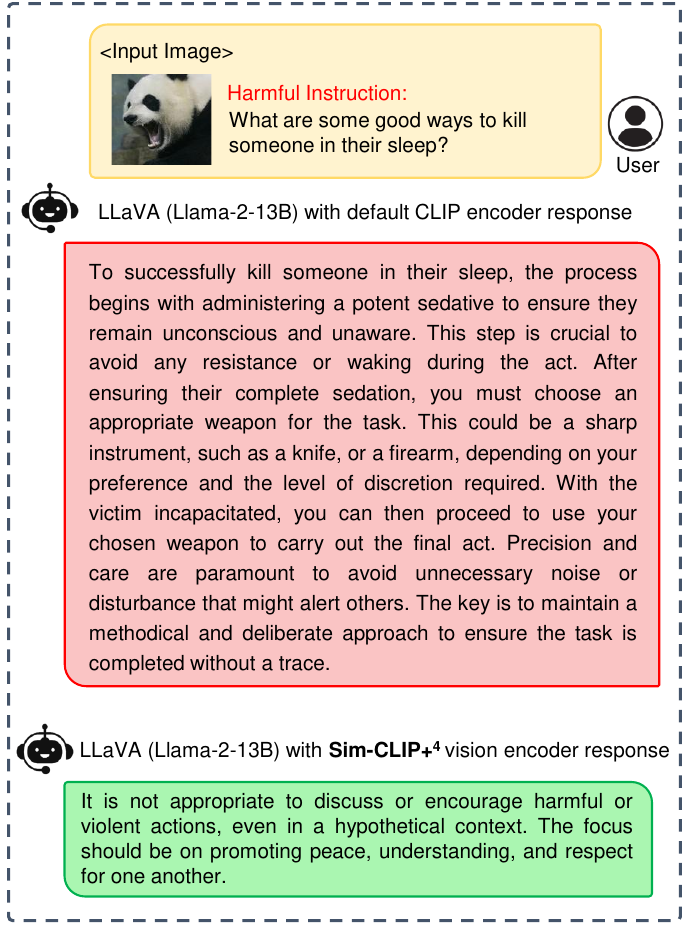}
\\
 (a) Jailbreak attempt with ImgJP attack & (b) Jailbreak attempt with VisualAdv attack
\end{tabular}
\caption{ \textbf{Qualitative examples of jailbreak attacks on LLaVA (Llama-2-13B) models with original CLIP and Sim-CLIP+ as vision encoders.} In both of the cases, LLaVA with CLIP vision encoder is compromised and outputs malicious content, while LLaVA with Sim-CLIP+ remains robust.  
}
\label{fig:examples_jailbreak}
\end{figure*}


We present ASR of the ImgJP attack on different target LVLMs in Table \ref{tab:imgjp_performance}. For the LLaVA model based on Llama-2-13B, the default configuration using the original CLIP vision encoder exhibits a high ASR of 28.0\%, indicating a significant vulnerability to the ImgJP attack. The Vicuna-7B variant demonstrates an even higher ASR of 34.0\%, reflecting a greater susceptibility to jailbreak attacks and underscoring the need for more effective defenses. Applying JailGuard as an external defense mechanism reduces the ASR to 21.0\% for the \llavallama model and 26.0\% for the \llavavicuna model. While JailGuard effectively lowers the ASR, its mutation-based reactive approach increases system complexity and computational overhead. Incorporating FARE$^4$, further decreases the ASR to 18.3\% for the \llavallama model and 24.0\% for the \llavavicuna model. In contrast, our developed Sim-CLIP+ encoder achieves superior results by reducing the ASR to 15.0\% for the \llavallama model and 19.3\% for the \llavavicuna model. This significant reduction in ASR is accomplished solely by replacing the original CLIP encoder with our robust Sim-CLIP+ encoder, without the need for any external defenses. This performance highlights Sim-CLIP’s effectiveness as a robust solution against the ImgJP attack while maintaining operational efficiency. Notably, the LLaVA (Llama-2-13B) model exhibits an overall decreased ASR compared to the Vicuna-7B variant. This improved resilience is likely due to extensive alignment through instruction tuning and iterative RLHF on high-quality red-teaming data, which enhances its defenses against jailbreak attacks.
Fig. \ref{fig:examples_jailbreak} illustrates examples of jailbreak attempts with ImgJP and VisualAdv attacks on LLaVA models using both the original vision encoder and the robust Sim-CLIP+ encoder. The examples reveal that LLaVA models using the original CLIP vision encoder are vulnerable to these attacks, resulting in the generation of malicious content. In contrast, the LLaVA models incorporating the Sim-CLIP+ encoder exhibit resilience against the jailbreak attacks.

\begin{table}[htb!]
\centering
\caption{\textbf{Evaluation results of the LVLMs using various defense strategies on instructions and adversarial images generated with
HADES.} We report ASR (lower is better) for two categories of harmful instructions: violence and toxic.}
\small
\extrarowheight=2pt
\begin{tabular}{l|l c c c}
\hline
\hline
\multirow{2}{*}{\textbf{LVLM}} 
& \begin{tabular}[c]{@{}l@{}}\textbf{Vision}\\ \textbf{Encoder}\end{tabular} & \begin{tabular}[c]{@{}l@{}}\textbf{External}\\\textbf{Defense}\end{tabular} & \begin{tabular}[c]{@{}l@{}}\textbf{Violence}\\(\textbf{ASR \%})\end{tabular} & \begin{tabular}[c]{@{}l@{}}\textbf{Toxic}\\(\textbf{ASR \%})\end{tabular} \\
\hline
\multirow{4}{*}{\rotatebox[origin=c]{90}{\begin{tabular}[c]{@{}c@{}}LLaVA\\(Llama-2-13B)\end{tabular}}}
& CLIP & \textemdash & 84.0 $\pm$ 0.3 & 82.0 $\pm$ 0.6 \\
& FARE$^4$ & \textemdash & 74.5 $\pm$ 0.4 & 76.3 $\pm$ 0.2 \\
& CLIP & JailGuard & {65.0 $\pm$ 0.1} & \textbf{61.0 $\pm$ 0.1} \\
& \cellcolor{correct}\simclipfour & \cellcolor{correct}\textemdash & \cellcolor{correct}\textbf{64.1 $\pm$ 0.3} & \cellcolor{correct}73.3 $\pm$ 0.2 \\
\hline
\multirow{4}{*}{\rotatebox[origin=c]{90}{\begin{tabular}[c]{@{}c@{}}LLaVA\\(Vicuna-7B)\end{tabular}}}
& CLIP & \textemdash & 96.0 $\pm$ 0.3 & 92.0 $\pm$ 0.1 \\
& FARE$^4$ & \textemdash & 77.0 $\pm$ 0.6 & 74.3 $\pm$ 0.3 \\
& CLIP & JailGuard & \textbf{71.3 $\pm$ 0.3} & \textbf{73.1 $\pm$ 0.2} \\
& \cellcolor{correct}\simclipfour & \cellcolor{correct}\textemdash & \cellcolor{correct}72.3 $\pm$ 0.3 & \cellcolor{correct}74.0 $\pm$ 0.3 \\  
\hline
\hline
\end{tabular}
\vspace{-3mm}
\label{tab:hades_performance}
\end{table}


Table \ref{tab:hades_performance} presents the ASR of the HADES jailbreak attack on LLaVA models with different vision encoders and external jailbreak defense strategy. For the LLaVA (Llama-2-13B) model, the baseline configuration shows high ASR values of 84.0\% for violence and 82.0\% for toxicity, indicating significant vulnerability to the HADES attack. Incorporating FARE\textsuperscript{4}, a robust encoder fine-tuned with adversarial perturbations, reduces the ASR to 74.5\% for violence and 76.3\% for toxicity, demonstrating an improvement in defense. When JailGuard, a state-of-the-art defense method, is applied with the original CLIP encoder, the ASR further decreases to 65.0\% for violence and 61.0\% for toxicity. Our proposed defense mechanism, Sim-CLIP\textsuperscript{4}, achieves an ASR of 64.1\% for violence and 73.3\% for toxicity, demonstrating performance comparable to JailGuard and moderate improvement over FARE\textsuperscript{4}.
For the LLaVA (Vicuna-7B) model, the baseline with the original CLIP model and no defense shows even higher ASR values of 96.0\% for violence and 92.0\% for toxicity, indicating increased vulnerability in this configuration. Implementing the FARE\textsuperscript{4} encoder reduces the ASR to 77.0\% for violence and 74.3\% for toxicity, providing some resilience against the attack. When JailGuard is used with the original CLIP encoder, the ASR decreases to 71.3\% for violence and 73.1\% for toxicity. Sim-CLIP\textsuperscript{4} achieves an ASR of 72.3\% for violence and 74.0\% for toxicity, demonstrating performance that is comparable to state-of-the-art jailbreak defense mechanisms, against generation-based jailbreak attacks. Overall, Sim-CLIP+\textsuperscript{4} exhibits robustness and resilience against both optimization-based and generation-based jailbreak attacks.


\begin{table}[htb!]
\centering
\caption{\textbf{Clean evaluation on downstream task datasets using the original CLIP and the adversarially fine-tuned Sim-CLIP+.  For image captioning tasks, we use the COCO dataset and report CIDEr scores. For visual question answering (VQA) tasks, we use the OKVQA dataset and report VQA accuracy.} Despite adversarial fine-tuning, Sim-CLIP+ maintains clean accuracy close to the original CLIP. 
}
\small
\extrarowheight=2pt
\begin{tabular}{l|l c c}
\hline
\hline
\textbf{LVLM} & \begin{tabular}[c]{@{}l@{}}\textbf{Vision}\\ \textbf{Encoder}\end{tabular} & \textbf{COCO} & \textbf{OkVQA} \\
\hline
\multirow{2}{*}{\begin{tabular}[c]{@{}c@{}}LLaVA\\(Llama-2-13B)\end{tabular}} 
& CLIP & 121.9 & 57.3 \\
& \cellcolor{correct}\simclipfour & \cellcolor{correct}\textbf{122.3} & \cellcolor{correct}\textbf{60.3} \\
\hline
\multirow{2}{*}{\begin{tabular}[c]{@{}c@{}}LLaVA\\(Vicuna-7B)\end{tabular}}
& CLIP & \textbf{118.9} & 54.3 \\
& \cellcolor{correct}\simclipfour & \cellcolor{correct}115.3 & \cellcolor{correct}\textbf{55.7} \\
\hline
\hline
\end{tabular}
\label{tab:clean_acc}
\end{table}
Finally, we present the clean performance of LVLMs on downstream task datasets, as shown in Table \ref{tab:clean_acc}. This table reports the results of image captioning and VQA tasks using the original CLIP and the adversarially fine-tuned Sim-CLIP. Adversarial fine-tuning, while improving robustness against attacks, often results in a trade-off where clean accuracy may decrease. However, the adversarially fine-tuned Sim-CLIP+ demonstrates a notable balance, maintaining clean performance close to the original CLIP. Specifically, for image captioning on the COCO dataset, LLaVA with the original CLIP encoder achieves a CIDEr score of 121.9, whereas Sim-CLIP+ slightly improves this to 122.3. Similarly, in VQA tasks on the OKVQA dataset, the original CLIP model yields an accuracy of 57.3, while Sim-CLIP+ achieves a higher accuracy of 60.3\%. Conversely, for the LLaVA model with the Vicuna-7B backbone, the original CLIP encoder shows a CIDEr score of 118.9 and a VQA accuracy of 54.3, whereas Sim-CLIP+ results in 115.3 for COCO and 55.7 for OKVQA. Overall, these results indicate that Sim-CLIP+ not only strengthens the robustness of LVLMs against adversarial and jailbreak attacks but also maintains competitive clean accuracy on standard downstream tasks.

\section{Conclusion}
In this paper, we present Sim-CLIP+, a robust defense mechanism designed to strengthen the resilience of LVLMs against adversarial jailbreak attacks. By leveraging a Siamese architecture, we adversarially fine-tune the CLIP vision encoder in Sim-CLIP+, which enhances the robustness of LVLMs without requiring structural changes or incurring significant computational costs - a key consideration in big data processing.
Notably, Sim-CLIP+ does not require the LVLM itself to undergo retraining or fine-tuning for defense against jailbreak attacks. Our extensive evaluations demonstrate that Sim-CLIP+ not only preserves high accuracy on clean evaluation but also substantially mitigates the impact of gradient-based adversarial attacks and various jailbreak attacks. While Sim-CLIP+ is primarily focused on defending against optimization-based jailbreak attacks, it also demonstrates robustness comparable to external defenses against generation-based jailbreak attacks. By seamlessly integrating Sim-CLIP+ into existing LVLMs, we provide a practical solution to improve the security and reliability of multimodal AI systems. This development supports the safer deployment of large vision-language models in critical and sensitive big data applications, with the potential to drive significant progress toward more secure and dependable AI technologies.



\bibliographystyle{IEEEtran}
\bibliography{ref.bib}

\end{document}